\documentclass[]{article}
\makeatletter\if@twocolumn\PassOptionsToPackage{switch}{lineno}\else\fi\makeatother

\usepackage{lineno}
\usepackage{amsfonts,amssymb,amsbsy,latexsym,amsmath,tabulary,graphicx, times,caption,fancyhdr, subcaption}
\usepackage[utf8]{inputenc}
\usepackage{cleveref}

\usepackage{url, multirow,morefloats,floatflt, cancel,tfrupee}
\makeatletter

\AtBeginDocument{\@ifpackageloaded{textcomp}{}{\usepackage{textcomp}}}
\makeatother
\usepackage{colortbl}
\usepackage{xcolor}
\usepackage{pifont}
\usepackage[nointegrals]{wasysym}
\usepackage{dsfont}
\makeatletter
\usepackage{numprint}

\def\mcWidth#1{\csname TY@F#1\endcsname+\tabcolsep}

\def\cAlignHack{\rightskip\@flushglue\leftskip\@flushglue\parindent\z@\parfillskip\z@skip}
\def\rAlignHack{\rightskip\z@skip\leftskip\@flushglue \parindent\z@\parfillskip\z@skip}

\@ifundefined{etal}{}{}

\usepackage{ifxetex}
\ifxetex\else\if@twocolumn\@ifpackageloaded{stfloats}{}{\usepackage{dblfloatfix}}\fi\fi

\AtBeginDocument{
\expandafter\ifx\csname eqalign\endcsname\relax
\def\eqalign#1{\null\vcenter{\def\\{\cr}\openup\jot\m@th
  \ialign{\strut$\displaystyle{##}$\hfil&$\displaystyle{{}##}$\hfil
      \crcr#1\crcr}}\,}
\fi
}

\AtBeginDocument{%
  \@ifpackageloaded{endfloat}%
   {\renewcommand\efloat@iwrite[1]{\immediate\expandafter\protected@write\csname efloat@post#1\endcsname{}}}{\newif\ifefloat@tables}%
}%

\def\BreakURLText#1{\@tfor\brk@tempa:=#1\do{\brk@tempa\hskip0pt}}
\let\lt=<
\let\gt=>
\def\processVert{\ifmmode|\else\textbar\fi}

\@ifundefined{subparagraph}{
\def\subparagraph{\@startsection{paragraph}{5}{2\parindent}{0ex plus 0.1ex minus 0.1ex}%
{0ex}{\normalfont\small\itshape}}%
}{}

\newcommand\role[1]{\unskip}
\newcommand\aucollab[1]{\unskip}

\@ifundefined{tsGraphicsScaleX}{\gdef\tsGraphicsScaleX{1}}{}
\@ifundefined{tsGraphicsScaleY}{\gdef\tsGraphicsScaleY{.9}}{}
\def\checkGraphicsWidth{\ifdim\Gin@nat@width>\linewidth
	\tsGraphicsScaleX\linewidth\else\Gin@nat@width\fi}

\def\checkGraphicsHeight{\ifdim\Gin@nat@height>.9\textheight
	\tsGraphicsScaleY\textheight\else\Gin@nat@height\fi}

\def\fixFloatSize#1{}%\@ifundefined{processdelayedfloats}{\setbox0=\hbox{\includegraphics{#1}}\ifnum\wd0<\columnwidth\relax\renewenvironment{figure*}{\begin{figure}}{\end{figure}}\fi}{}}
\let\ts@includegraphics\includegraphics

\def\inlinegraphic[#1]#2{{\edef\@tempa{#1}\edef\baseline@shift{\ifx\@tempa\@empty0\else#1\fi}\edef\tempZ{\the\numexpr(\numexpr(\baseline@shift*\f@size/100))}\protect\raisebox{\tempZ pt}{\ts@includegraphics{#2}}}}

\AtBeginDocument{\def\includegraphics{\@ifnextchar[{\ts@includegraphics}{\ts@includegraphics[width=\checkGraphicsWidth,height=\checkGraphicsHeight,keepaspectratio]}}}

\DeclareMathAlphabet{\mathpzc}{OT1}{pzc}{m}{it}

\def\URL#1#2{\@ifundefined{href}{#2}{\href{#1}{#2}}}

\def\UrlOrds{\do\*\do\-\do\~\do\'\do\"\do\-}%
\g@addto@macro{\UrlBreaks}{\UrlOrds}

\DeclareMathOperator*{\argmax}{arg\,max}

\edef\fntEncoding{\f@encoding}

\makeatother

\newif\ifmultipleabstract\multipleabstractfalse%
\usepackage{algorithm}
\usepackage{algpseudocode}

\makeatletter

\def\wileyIndent{1pt}
\usepackage[paperheight=10in,paperwidth=6.5in,margin=2cm,headsep=.5cm,top=2.5cm,headheight=1cm]{geometry}

\renewenvironment{abstract}
{\vspace*{-1pc}\trivlist\item[]\leftskip\wileyIndent\par\vskip4pt\noindent\textbf{\abstractname}\mbox{\null}\\}{\par\noindent\endtrivlist}

\usepackage[]{footmisc}

\def\title#1{\linespread{1}\gdef\@title{\centering\bfseries\ifx\@articleType\@empty\else\@articleType\\\fi#1}}

\let\@articleType\@empty \def\articletype#1{\gdef\@articleType{{\normalfont\itshape#1}}}

 \def\audegree#1{}

\Crefformat{figure}{#2Fig.~#1#3}
\newcommand*\bigcdot{\mathpalette\bigcdot@{.75}}
\newcommand*\bigcdot@[2]{\mathbin{\vcenter{\hbox{\scalebox{#2}{$\m@th#1\bullet$}}}}}
\date{}

\makeatother

\usepackage[T1]{fontenc}
\makeatother
\usepackage[style=authoryear, natbib=true, backend=biber, giveninits=true]{biblatex}
\usepackage[textwidth=20mm]{todonotes}

\begin{document}

\title{\vspace{-2cm} Cooperative learning of Pl@ntNet's Artificial Intelligence algorithm: how does it work and how can we improve it?}

\author{
Tanguy Lefort $^{1}$\and Antoine Affouard $^{2}$\and Benjamin Charlier $^{3}$\and Jean-Christophe Lombardo $^{4}$\and Mathias Chouet $^{5}$\and Hervé Goëau$^{6}$\and Joseph Salmon $^{7}$ \and Pierre Bonnet $^{8}$\and Alexis Joly $^{9}$ \\
    $^{1}$\small{Univ. Montpellier, CNRS, IMAG, Inria, LIRMM, France tanguy.lefort@umontpellier.fr}\\
    $^{2}$ \small{Inria, CIRAD, Montpellier, France antoine.affouard@cirad.fr }\\
    $^{3}$ \small{Univ. Montpellier, CNRS, IMAG, France benjamin.charlier@umontpellier.fr }\\
    $^{4}$ \small{Inria, LIRMM, France, jean-christophe.lombardo@inria.fr}\\
    $^{5}$ \small{CIRAD, AMAP, Montpellier, France, mathias.chouet@cirad.fr}\\
    $^{6}$ \small{CIRAD, AMAP, Montpellier, France, herve.goeau@cirad.fr}\\
    $^{7}$ \small{Univ. Montpellier, CNRS, IMAG, Institut Universitaire de France (IUF), joseph.salmon@umontpellier.fr} \\
    $^{8}$ \small{CIRAD, AMAP, Montpellier, France, pierre.bonnet@cirad.fr}\\
    $^{9}$ \small{Inria, LIRMM, France, alexis.joly@inria.fr}\\
}

\maketitle
\thispagestyle{empty} 
\paragraph*{Running Headline}

Citizen science for plant identification
% Pl@ntNet: from citizen to AI identifications

\paragraph*{Acknowledgments}
This work was funded by the French National Research Agency (ANR) through the grant Pl@ntAgroEco 22-PEAE0009, granted access to the HPC resources of IDRIS under the allocation A0151011389 made by GENCI, and funded by the Chaire IA CaMeLOt (ANR-20-CHIA-0001-01).

\paragraph*{Data Availability}
The dataset is available at \url{https://doi.org/10.5281/zenodo.10782465}

\paragraph*{Conflict of Interest}
The authors declare no conflicts of interest

\paragraph*{Author Contributions}
T. Lefort, A. Affouard, A. Joly, B. Charlier, P. Bonnet and J. Salmon conceived the ideas and designed the evaluation methodology; A. Affouard and M. Chouet are the main developers of Pl@ntNet's backend ; T. Lefort and A. Affouard collected the evaluation data used in this paper; T. Lefort re-implemented Pl@ntNet's algorithm in python and conducted the evaluation ; J-C. Lombardo, H. Goëau and A. Joly conceived and trained Pl@ntNet's AI model; T. Lefort, B. Charlier, A. Joly and J. Salmon analyzed the outcomes of the study. All authors contributed critically to the drafts and gave final approval for publication.

\clearpage
\setcounter{page}{1}
\begin{abstract}
    \begin{enumerate}
        \item
        Deep learning models for plant species identification rely on large annotated datasets. The Pl@ntNet system enables global data collection by allowing users to upload and annotate plant observations, leading to noisy labels due to diverse user skills. Achieving consensus is crucial for training, but the vast scale of collected data (number of observations, users and species) makes traditional label aggregation strategies challenging. Existing methods either retain all observations, resulting in noisy training data or selectively keep those with sufficient votes, discarding valuable information. Additionally, as many species are rarely observed, user expertise can not be evaluated as an inter-user agreement: otherwise, botanical experts would have a lower weight in the AI training step than the average user.
        \item
        Our proposed label aggregation strategy aims to cooperatively train plant identification AI models. This strategy estimates user expertise as a trust score per user based on their ability to identify plant species from crowdsourced data. The trust score is recursively estimated from correctly identified species given the current estimated labels. This interpretable score exploits botanical experts’ knowledge and the heterogeneity of users. Subsequently, our strategy removes unreliable observations but retains those with limited trusted annotations, unlike other approaches.
        \item
        We evaluate Pl@ntNet's strategy on a newly released large subset of the Pl@ntNet database focused on European flora, comprising over 6M observations and 800K users. This anonymized dataset of votes and observations is released openly at {\url{https://doi.org/10.5281/zenodo.10782465}}. We demonstrate that estimating users’ skills based on the diversity of their expertise enhances labeling performance.
        \item
        Our findings emphasize the synergy of human annotation and data filtering in improving AI performance for a refined training dataset. We explore incorporating AI-based votes alongside human input in the label aggregation. This can further enhance human-AI interactions to detect unreliable observations (even with few votes).

    \end{enumerate}
\def\keywordstitle{Keywords}
\keywordstitle: crowdsourcing, botanical skills, human-AI interaction, label aggregation, Pl@ntNet, plant identification
\end{abstract}

\newpage

\section{Introduction}

Computer vision models are a great aid in plant species recognition in the field \citep{vidal2021perspectives,borowiec2022,mader2021flora}.
However, to train them we need large annotated datasets.
These datasets are often created thanks to citizen science approaches, collecting both reliable and useful information \citep{brown2019potential}.
Among existing plant recognition applications, the Pl@ntNet citizen science platform \citep{affouard2017pl} enables global data collection by allowing users to upload and annotate plant observations \citep{bonnet2020citizen}.

\begin{figure}[htb]
    \centering
    \includegraphics[width=.75\linewidth]{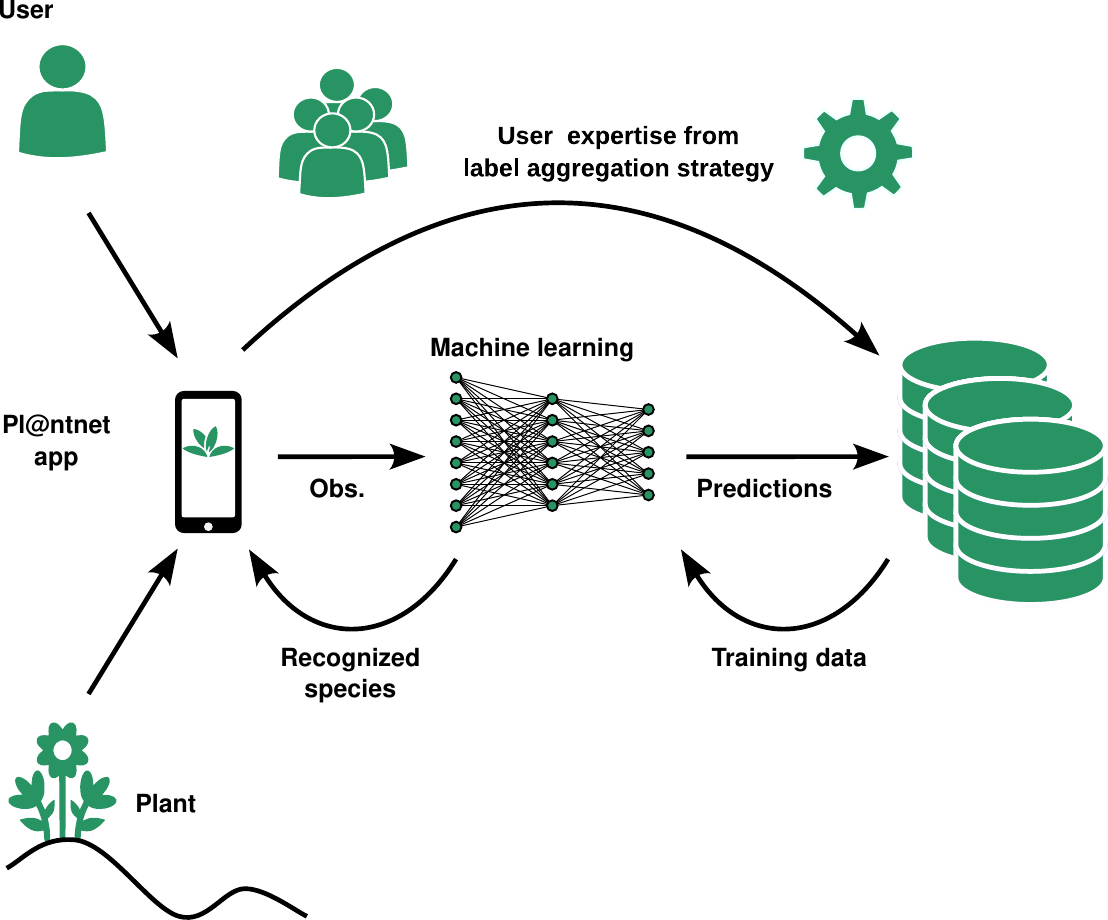}
    \caption{Pl@ntNet system of human-AI interaction for plant species recognition. Users take their plant observations in the Pl@ntNet application. A prediction is output by the AI model. Users can validate the prediction or propose another species. The whole votes collection is used to evaluate user expertise (see \Cref{alg:plantnet_algorithm}) and actively revise observations identifications.}
    \label{fig:plantnet-system}
\end{figure}

At the time of writing, this participatory approach has resulted in the collection of over 20 million observations (images or group of images of the same plant), belonging to almost $\numprint{46000}$ species, by more than 6 million users worldwide. In total, more than $25$ million of images are shared in these observations. The collaborative process of Pl@ntNet is synthetized in \Cref{fig:plantnet-system}.
The AI model interacts with the human decision by proposing possible species given an observation.
For each returned species, using a similarity search, the Pl@ntNet system also shows similar pictures from the database.
This lets users visually check that their observation is likely to belong to a predicted species given the most similar observations.
For instance, such a visual control can help to compare two plants at various growth stages.

Plant species identification is a task that requires skills to recognize morphological traits (shapes, measurements, environments and specific characteristics).
A large number of users with diverse skills have participated in gathering plant observations and helped improve the training dataset of our computer vision model.
Their participation is based on votes that they can cast on others' observations, or by the initial species determination of their observation.
% Pl@ntNet citizen data collection allows gathering a large number of observations but also a large number of users with different skills.
The quality of each vote is then processed by the algorithm presented in \Cref{subsec:label_aggregation_strategy}.

Other citizen science projects such as iNaturalist \citep{van2018inaturalist} or eBird \citep{sullivan2009ebird} use a similar approach to collect data, but differ in their label aggregation strategy.
The iNaturalist project, with more than $2.5$ million users, records the votes at different taxonomic levels.
The resulting label is the aggregation of at least two votes on a species-level identification (or coarser or finer taxonomic level).
A taxon requires at least a two-thirds agreements among identifiers and all users have the same weight in the decision-making.
Over time, a taxon can be further refined by the community, debated or revoked.
eBird handles taxon quality control by using a checklist in each region for observers.
Quality control on the checklist is performed and, combined with user knowledge -- number of species and checklist submitted, number of flagged observations, discussions among local experts -- the species observation is accepted.
The eBird project also showed that monitoring species accumulation from observers can help to sort their skills \citep{kelling2015}. While they consider the species accumulation by hours spent on each collected observation, we propose a strategy that takes into account the entire history of observations of the observer.

% Most reliable Pl@ntNet observations have been shared with the Global Biodiversity Information Facility (GBIF).
% This allows researchers to have a map of the presence of observed plant species anywhere on Earth.
% This is a biased indicator of the true state of the ecosystem, as the citizen-based data collection approach is not random.
% There are both spatial and temporal biases as users are more likely to observe plants in their environment and during their free time as reported in \citet{di2021observing} and \citet{courter2013weekend}.

In this article, we present the Pl@ntNet label aggregation strategy.
Using a new large-scale dataset of more than $6$ million observations and $800$ thousand users, we show that our strategy can improve the quality of the collected data, without removing every observation that was only labeled by single users.
Finally, aggregated labels are used in practice to train an AI model. We explore how the information contained in the AI predictions can be integrated into the label aggregation strategy to generate new votes and help control data quality.
By using the model's predictions within the label aggregation, the goal is to correct possible mistakes from non expert users without contradicting botanical experts. 
% We also show that our strategy can be used to train an AI model that can generate new votes and help control data quality.

\section{Methods}

%%%%%%%%%%%%%%%%%%%%%%%%%%%%%%%%%%%%%%%%%%%%%%%%%%%%%%%%%%%%%%%%%%%%%%%%%%%%%%%
\subsection{Dataset and notation}
\label{subsec:dataset}
%%%%%%%%%%%%%%%%%%%%%%%%%%%%%%%%%%%%%%%%%%%%%%%%%%%%%%%%%%%%%%%%%%%%%%%%%%%%%%%

To compare the different label aggregation strategies on large-scale datasets, we introduce a subset of the Pl@ntNet database focused on Southwestern European flora observations -- Baleares, Corsica, France, Portugal, Sardegna and Spain -- from $2017$ to October $2023$.
In total, $\numprint{9005108}$ votes are cast by $n_\text{user}=\numprint{823251}$ users on $\numprint{6699593}$ observations after two cleaning steps on the voted species.
The first one is a filtering step. We only keep the votes with plant species belonging to the World Checklist of Vascular Plants (WCVP) \citep{govaerts2023world}.
For the second step, according to Kew's Royal Botanical Garden, we matched synonyms to their backbone species if the species is part of the \emph{k-southwestern-europe} checklist from Plants of the World Online \citep{powo2024} (POWO) system.
% For the second step, we handled synonyms according to Kew's Royal Botanical Garden checklist.
% We restrict synonyms to be part of the \emph{k-southwestern-europe} from Plants of the World Online \citep{powo2024} (POWO) system.
Note that there are plant species listed in the accepted species from WCVP that are not in the \emph{k-southwestern-europe} POWO checklist.
As there is a possible taxon ambiguity in this case -- multiple species possible for a given synonym depending on the referential -- we leave the proposed label untouched.
The dataset is available at \url{https://zenodo.org/records/10782465}.
% Note that however, plant species proposed in the WCVP but not in the \emph{k-southwestern-europe} POWO reference do not have a synonym filter.

%Among the users, $98$ are identified as botanical experts by the Pl@ntNet team and Telabotanica platform. The answers of these experts are considered ground truth labels and used in the evaluation step. Observations where multiple experts disagree are removed from the test set.
%The test set is finally composed of $26,811$ observations with $17,125$ with at least two identification votes.
%\Cref{fig:sankey} shows the distribution of observations in the dataset.

\paragraph*{Notation}

In the following, denote $K$ the number of species within the dataset.
We index the observations by $i\in [n_{\bigcdot}]=\{1,\dots,n_{\bigcdot}\}$ where $\mathcal{D}_{\bigcdot}$ is the considered dataset composed of $n_{\bigcdot}$ observations and their associated votes.
For example, the full south-western European flora dataset from Pl@ntNet of $n_{\mathrm{SWE}}=\numprint{6699593}$ observations is denoted $\mathcal{D}_{\mathrm{SWE}}$.
Other subsets are presented in \Cref{subsec:existing_label_aggreagation_strategies}.
We write $\mathcal{U}$ the set of all users. Each user $u$ has a unique identifier used as an index, and we denote $\mathcal{U}_i$ the set of users that have voted on observation $i$ -- \emph{i.e.} $\mathcal{U} = \cup_{i\in [n_\mathrm{SWE}]} \mathcal{U}_i$.
The vote of user $u$ on observation $i$ is denoted $y_i^u\in [K]$.
Estimated labels are denoted $\hat y_i \in [K]$.
Each observation $i$ is created by an author $u$ stored in $\mathrm{Author}(i)$.

%%%%%%%%%%%%%%%%%%%%%%%%%%%%%%%%%%%%%%%%%%%%%%%%%%%%%%%%%%%%%%%%%%%%%%%%%%%%%%%
\subsection{Proposed label aggregation strategy}
\label{subsec:label_aggregation_strategy}
%%%%%%%%%%%%%%%%%%%%%%%%%%%%%%%%%%%%%%%%%%%%%%%%%%%%%%%%%%%%%%%%%%%%%%%%%%%%%%%

\begin{algorithm}[H]
    \caption{Pl@ntNet iterative weighted majority vote}
    \label{alg:plantnet_algorithm}
    \begin{algorithmic}[1]
        \Require Votes as $(u, y_i^u)_{i\in [n_{\mathrm{SWE}}],u\in [n_\text{user}]}$ for each observation $i$ and user $u$ answering the voted species $y_i^u$, accuracy threshold $\theta_{\text{acc}}$, confidence threshold $\theta_{\text{conf}}$, weight function $f$, initial weight $\gamma>0$
        \Ensure Estimated labels $\hat y_i$ and validity indicator $s_i$ for each observation $i$
        % \State $\text{Initialize } \hat y_i = \mathrm{MV}\left(\{y_i^u\}_u\right) \text{ for each observation } i \in [n_{\mathrm{SWE}}]$
        \State $\text{Initialize user weights as } w_u = \gamma \text{ for each user } u \in [n_\text{user}]$
        \While{$\text{not converged}$}
        \State Get current estimated labels with a weighted majority vote $$ \forall i \in [n_{\mathrm{SWE}}],\ \hat y_i = \argmax_{k\in[K]} \sum_{u\in \mathcal{U}_i} w_u\mathds{1}(y_i^u=k)$$
            \For{each observation $i\in [n_{\mathrm{SWE}}]$}
                \State Compute label confidence: $\mathrm{conf}_i(\hat y_i) = \sum_{u\in \mathcal{U}_i} w_u \mathds{1}(y_i^u=\hat y_i)$
                \State Compute label accuracy: $\mathrm{acc}_i(\hat y_i) = \mathrm{conf}_i(\hat y_i) / \sum_{k\in [K]} \mathrm{conf}_i(k)$
                \State Compute validity indicator: $s_i = \mathds{1}( \mathrm{acc}_i(\hat y_i) \geq \theta_{\text{acc}} \text{ and } \mathrm{conf}_i(\hat y_i) \geq \theta_{\text{conf}})$
            \EndFor
            \For{each user $u \in [n_\text{user}]$}
                \State Compute the number of valid identified species for authoring observations: \[n_u^\text{author} = |\{y_i^u\in [K] \,|\, y_i^u=\hat y_i, s_i=1, \mathrm{Author}(i)=u\}|\]
                \State Compute the number of identified species by voting on other's observations: \[n_u^\text{vote}=|\{y_i^u\in [K]\,|\, y_i^u=\hat y_i, \mathrm{Author}(i)\neq u\}|\]
                \State Compute the rounding number of identified species per user: \[n_u = \mathrm{Round}\left(n_u^\text{author} + \frac{1}{10}n_u^\text{vote}\right)\]
                \State Transform number of estimated species per user into trust score: $w_u = f(n_u)$
                \EndFor
        \EndWhile
    \end{algorithmic}
\end{algorithm}

Pl@ntNet label aggregation strategy relies on estimating the number of correctly identified species for each user.
Similar to other strategies, we rely on an EM-based iterative procedure \citep{Dempster_Laird_Rubin77} to estimate consecutively the users' skills and each observation's species. The detailed iterative algorithm is provided in \Cref{alg:plantnet_algorithm} and available at \url{https://github.com/peerannot/peerannot}. 
% As the collected data is used to train the AI model,
The label aggregation strategy generates a trust indicator ($s_i$) on the observation that can reveal whether an observation is valid or not.
Notice that in \Cref{alg:plantnet_algorithm} we value $10$ times more authored observations than voting on other's observations -- if a user proposes a new observation with a label (species name) it is more useful than proposing a label by clicking.
Indeed, being on the field leads to more information on the environment and a better determination of the species.
Finally, note that an identified species is exclusively identified as author -- part of $n_u^\text{author}$ in \Cref{alg:plantnet_algorithm}) -- or as click -- part of $n_u^\text{vote}$ -- to avoid redundant skills.
%species are unequivocally identified as authors ($n_u^\text{author}$ in \Cref{alg:plantnet_algorithm}) or as votes on other's observations ($n_u^\text{vote}$).
The final number of species identified by users is the aggregation of these two terms: $n_u = \mathrm{Round}\left(n_u^\text{author} + \frac{1}{10}n_u^\text{vote}\right)$.

\begin{figure}[tbh]
    % \centering
    % \includegraphics[width=.8\textwidth]{images/weight_function.pdf}
    % \caption{Weight function (\Cref{eq:weight_func}) used to map the number of identified species to a trust score in the Pl@ntNet label aggregation strategy. The user confidence threshold $\theta_{\text{conf}}=2$ requires a user to have identified at least $n_u=8$ species to become \textbf{self-validating}. A new user starts with a weight of $f(0)=f(1)=\gamma\simeq 0.74$.}
    % \begin{subfigure}[t]{.48\textwidth}
        \centering
        \includegraphics[width=.8\textwidth]{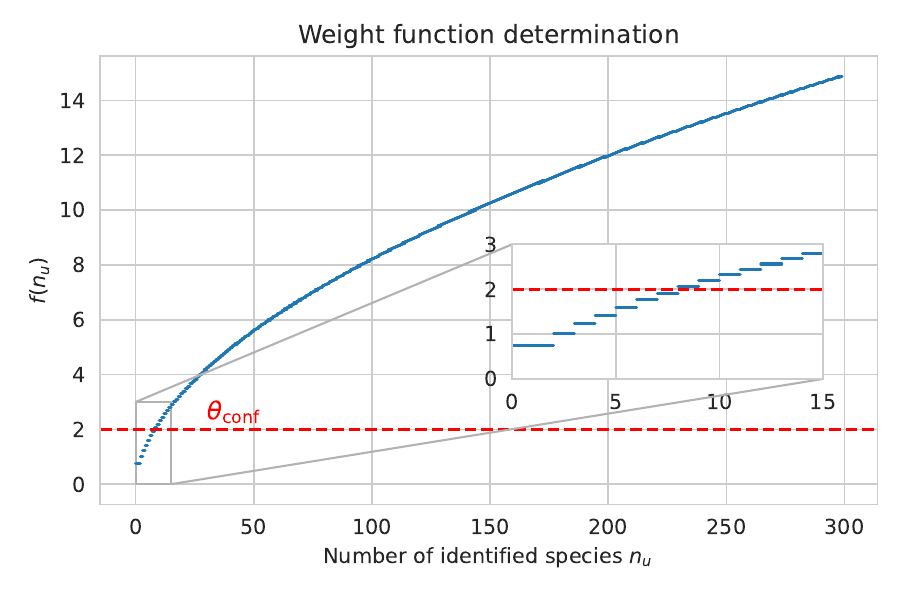}
        % \caption{Weight function in \Cref{eq:weight_func} used to map the number of identified species to a trust score in the Pl@ntNet label aggregation strategy.  A new user starts with a weight of $f(0)=f(1)=\gamma\simeq 0.74$.}
        % \label{fig:weight_function}
    % \end{subfigure}%
    % \hfill
    % \begin{subfigure}[t]{.48\textwidth}
    %     \centering
    %     \includegraphics[width=\linewidth]{./images/weight_function_exploration.pdf}
    %     \caption{Visualization of the two regimes in Pl@ntNet's weight function. The first regime separates users with low participation. The second regime imitates the square root function -- as $\alpha=0.5$ -- to separate expert users.}
    %     \label{fig:weight_exploration}
    % \end{subfigure}
    \caption{Weight function in \Cref{eq:weight_func} used to map the number of identified species to a trust score in the Pl@ntNet label aggregation strategy. A new user starts with a weight of $f(0)=f(1)=\gamma\simeq 0.74$. The user confidence threshold $\theta_{\text{conf}}=2$ requires a user to have identified at least $n_u=8$ species to become \textbf{self-validating}. The parameters $\alpha=0.5$, $\beta=0.2$ and $\gamma\simeq 0.74$ are used in practice.}
    \label{fig:weight_function}
\end{figure}

The weight function $f$ shown in \Cref{fig:weight_function} is a non-decreasing function that maps the number of identified species $n_u$ to a trust score in the form of:
\begin{equation}\label{eq:weight_func}
    w_u = f(n_u) = n_u^\alpha - n_u^\beta + \gamma \enspace,
    \end{equation}
where $\alpha,\beta\in\mathbb{R_+^\star}$ are hyperparameters that were calibrated internally to fit prior knowledge and $\gamma>0$ is the constant representing the initial weight of each user. In practice, we use $\alpha=0.5$, $\beta=0.2$ and $\gamma=\log(2.1)\simeq 0.74$ in the weight function.
This function is sub-linear ($\mathcal{O}(\sqrt{n_u}))$ but with two different behaviors.
The goal of \Cref{eq:weight_func} is to separate new users from experts and then help sort multiple experts.
This is modeled by the two behaviors of the weight function.
In the first part which corresponds to new users with low $n_u$, the term in the power of $\beta$ decreases the weight. We chose an initial weight $w_u=\gamma$ such that a user has a weight equal to $1$ (rounding to two decimals) with two different identifications. This separates the users who only come once to test the application from others.
In the second part with a higher number of identified species, the term to the power of $\beta$ becomes negligible and we tend to the square root function.
The sub-linear scale allows for reducing discrepancies between people who have identified a comparable number of species (and thus have presumably comparable expertise).
As for the two thresholds that control the level of uncertainty accepted for a given label, they are set to $\theta_{\text{conf}}=2$ to control the total weight on an observation and $\theta_{\text{acc}}=0.7$ to control the agreement between users given their expertise.

Users are said \textbf{self-validating} when they are trusted enough so that their proposed label single-handedly makes an observation valid $(s_i=1)$.
From \Cref{alg:plantnet_algorithm}, we see that this is verified when their weight $w_u$ is greater than the level $\theta_{\text{conf}}$.
Indeed, with a single label we obtain $\mathrm{conf}_i(\hat y_i) = w_u > \theta_{\text{conf}}$ and $\mathrm{acc}_i(\hat y_i) = 1 > \theta_{\text{acc}}$.
In practice, this means that an experienced user who has collected enough weight can validate any observation without any other user's vote.
Note that this identification can later be invalidated by other users with enough weight thanks to the accuracy threshold $\theta_{\text{acc}}$.

%%%%%%%%%%%%%%%%%%%%%%%%%%%%%%%%%%%%%%%%%%%%%%%%%%%%%%%%%%%%%%%%%%%%%%%%%%%%%%%
\subsection{Evaluation against other aggregation strategies}
\label{subsec:existing_label_aggreagation_strategies}
%%%%%%%%%%%%%%%%%%%%%%%%%%%%%%%%%%%%%%%%%%%%%%%%%%%%%%%%%%%%%%%%%%%%%%%%%%%%%%%

\paragraph*{Existing aggregation strategies.}

Plant species label aggregation is a challenging task due to the large number of species $K=\numprint{11425}$.
Hence, many classical strategies in the label aggregation literature such as Dawid and Skene's \citep{dawid_maximum_1979} and other variations \citep{passonneau-carpenter-2014-benefits, sinha2018fast} are not applicable as they require estimating a $K\times K$ confusion matrix for each user.
For the considered dataset $\mathcal{D}_\text{SWE}$, this would result in $\numprint{11425}^2\times \numprint{823251}\approx 10^{14}$ parameters to estimate.
Similar issues occur for other label aggregation strategies \citep{whitehill_whose_2009,hovy2013learning,ma2020adversarial}.
We do not consider deep-learning-based crowdsourcing strategies as \citet{rodrigues2018deep,chu2021learning} or \citet{lefort2022improve} as they require training a neural network from crowdsourced labels, but do not output aggregated labels on the training set.
In the Pl@ntNet application, we need to propose one or multiple species for each observation to users.
To overcome these issues, we consider the following label aggregation strategies that can scale with $K$ and the number of users:
\begin{itemize}
    \item \textbf{Majority Vote (MV)} \citep{james1998majority}: it selects the most answered label\footnote{Ties are broken at a random -- creating sometimes some variability in the labeling process.} and is the most common aggregation strategy. More formally, given an observation $i$:
    \[\mathrm{MV}(i, \{y_i^u\}_u) = \argmax_{k\in [K]} \sum_{u\in \mathcal{U}_i} \mathds{1}(y_i^u = k) \enspace.\]
    \item \textbf{Worker Agreement With Aggregate (WAWA)} \citep{appen_wawa_2021}: this strategy, also known as the inter-rater agreement, weights each user by how much they agree with the MV labels on average. More formally, given an observation $i$:
    \begin{align*}
        \mathrm{WAWA}(i, \mathcal{D}_{\mathrm{SWE}}) &= \argmax_{k\in [K]} \sum_{u\in \mathcal{U}_i} w_u\mathds{1}(y_i^u = k) \\
        \text{with } w_u &= \frac{1}{|\{y_{i'}^u\}_{i'}|} \sum_{i'=1}^{n_{\mathrm{SWE}}} \mathds{1}\left(y_{i'}^u = \mathrm{MV}(i',\{y_{i'}^u\}_u)\right)\enspace.
        \end{align*}
    As there is no observation filter for the $\mathrm{MV}$ and $\mathrm{WAWA}$, we consider that for all observation $i$, $s_i=1$ for these two strategies.

    \item \textbf{TwoThird}: The TwoThird aggregation generates a label for observations with at least two votes. The estimated label represents the one with at least two-thirds of the majority in agreement. Every user has the same weight in the aggregation. It is part of the iNaturalist's label aggregation system \citep{van2018inaturalist}. More formally:
    \begin{align*}
        \mathrm{TwoThird}(i, \{y_i^u\}_u) &= \begin{cases}
            \mathrm{MV}(i, \{y_i^u\}_u) & \text{if } s_i=1  \\
            \text{undefined} & \text{otherwise} \end{cases}
            \\ \text{ with } s_i &= \mathds{1}\left(\displaystyle\max_{k\in[K]}\frac{1}{|\mathcal{U}_i|}\sum_{u\in \mathcal{U}_i} \mathds{1}(y_i^u=k) \geq \frac{2}{3}\right) \enspace.
    \end{align*}
\end{itemize}

\paragraph*{Creation of an evaluation set in a crowdsourcing setting.}

To evaluate the performance of a label aggregation strategy, it is necessary to know the ground truth on a subset of the data.
However, in the context of crowdsourced data, there is no known truth for the observations.
The sheer volume of data makes it impossible to ask botanical experts to create such ground truth for the whole database.
Moreover, identifying species from images is much less accurate than identifying them in the field, due to the partial information contained in the image \citep{experts2018plant}. 

Instead of asking experts to label a subset of the data, we rather identify botanical experts in the Pl@ntNet user database and consider their determinations as ground truth.
We asked botanical-known experts to reference other experts who could have a Pl@ntNet account to create a list of expert users.
To these we have added TelaBotanica \citep{heaton2010tela} users with registered confirmed botanical experience from their account and that are also Pl@ntNet users that participated in the South-Western Europe flora subset.
In total, $98$ Pl@ntNet users were identified as botanical experts.
Observation with at least one vote from one of these experts constitute our test set denoted $\mathcal{D}_\text{expert}$.
The answers of these experts are considered ground truth labels and used to evaluate strategies' performance.
Despite our selection process of supposedly `indisputable' experts, a few observations in the test set still end up with contradictory labels ($4$ observations in total). As they represent a very small percentage, we simply removed them from $\mathcal{D}_\text{expert}$.

\begin{figure}[tbh]
    \begin{center}
        \includegraphics[width=\textwidth]{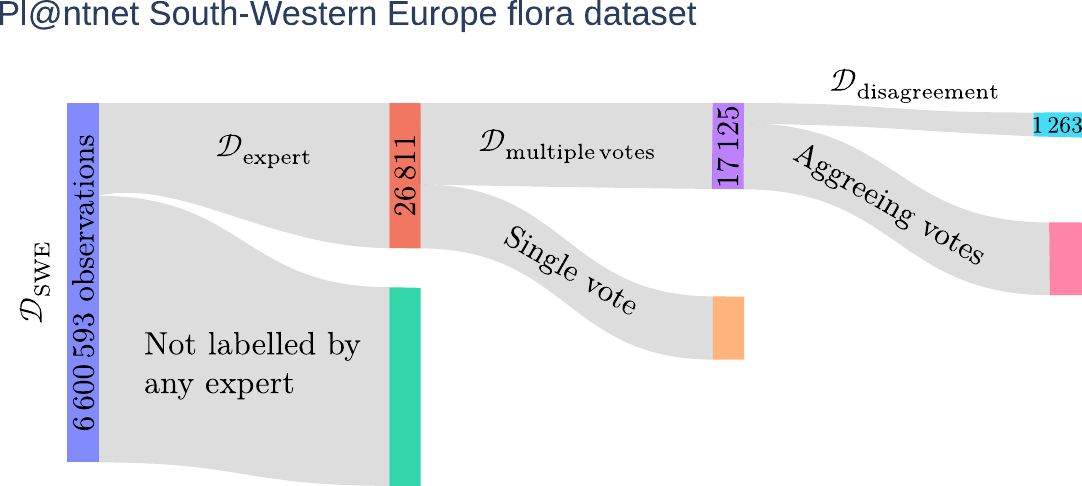}
    \end{center}
    \caption{Log-scales distribution of the observations in the South-West European Flora subset from the Pl@ntNet database. Note that the (sub-)datasets introduced are nested: $\mathcal{D}_{\mathrm{SWE}} \supset \mathcal{D}_\text{expert} \supset \mathcal{D}_{\text{multiple votes}} \supset \mathcal{D}_{\text{disagreement}}$. $\mathcal{D}_{\text{expert}}$ and the following subsets contain observations that received at least one vote from one of the experts.}
    \label{fig:sankey}
\end{figure}

\begin{figure}[tbh]
    \centering
    \begin{subfigure}[t]{.52\textwidth}
        \centering
        \includegraphics[width=\linewidth]{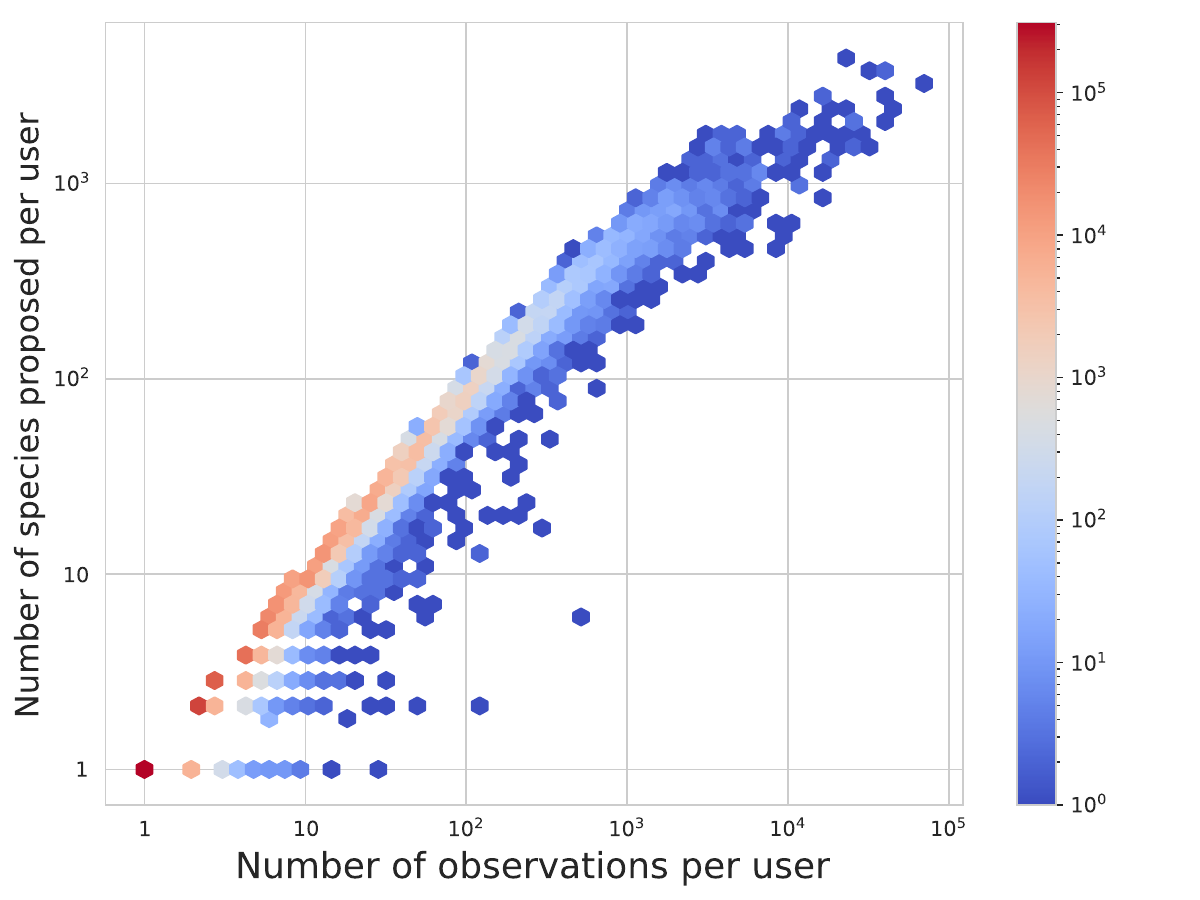}
        \caption{Relationship between the number of observations per user and the variety of species proposed per user. Each point represents a concentration of users in the SWE flora subset. $\numprint{310,564}$ users proposed a single vote.}
        \label{fig:users_obs_species}
    \end{subfigure}%
    \hfill
    \begin{subfigure}[t]{.40\textwidth}
        \centering
        \includegraphics[width=\linewidth]{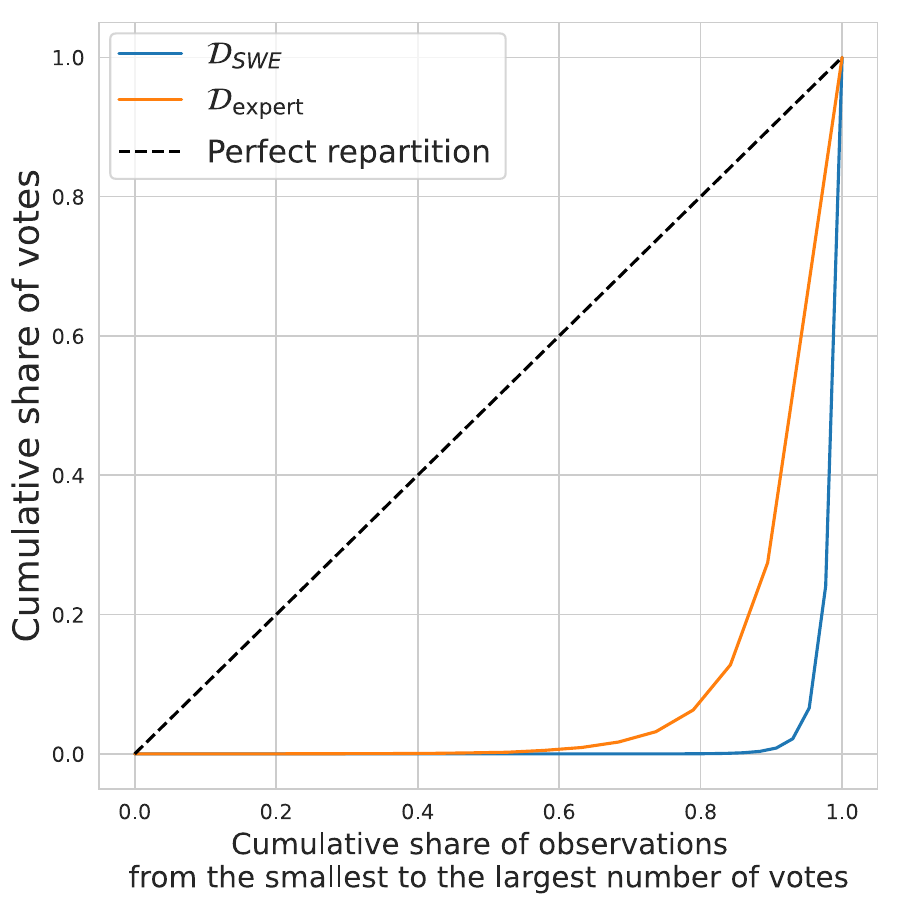}
        \caption{Lorenz curves representing the imbalance distribution of the number of votes in the South-West European Flora subset from the Pl@ntNet database. This imbalance is mitigated but kept in the created test set.}
        \label{fig:lorenz}
    \end{subfigure}
    \caption{Pl@ntNet activity summary in the SWE flora subset. (A): The majority of users have proposed a small number of observations and species. However, some users have proposed a large number of observations and species. (B): In a perfectly balanced dataset, the Lorenz curve would be the diagonal -- $50\%$ of the votes would be for $50\%$ of the observations. In practice, there is a high imbalance of the distribution of votes between observations -- $80\%$ of the observations are represented by $10\%$ of votes.}
    \label{fig:summary_subset}
\end{figure}

Our evaluation set $\mathcal{D}_{\text{expert}}$ is finally composed of $\numprint{26811}$ observations which received at least one vote from one of the experts.
Despite the large number of users, not all observations obtain multiple annotations.
Indeed, $\numprint{310564}$ users were single-time voters (meaning they interacted with the system only once).
The lack of votes is a large component of difficulty in the Pl@ntNet database, as there is a high imbalance of the distribution of votes between observations as represented in \Cref{fig:lorenz}.
There is a high concentration of votes for a small percentage of the observations as shown in \Cref{fig:users_obs_species}.
Of these evaluation data, $\numprint{17125}$ received more than two identifications and are stored in $\mathcal{D}_{\text{multiple votes}}$.
Then, $\numprint{1263}$ have more than two votes with at least one disagreement between users and are stored in $\mathcal{D}_{\text{disagreement}}$.
\Cref{fig:sankey} shows the distribution of observations from $\mathcal{D}_{\mathrm{SWE}}$ to the finer and more ambiguous $\mathcal{D}_{\text{disagreement}}$.

% Unfortunately, the demand for multiple labels on observations is not being met, despite the large number of users.

\paragraph*{Evaluation metric.}

To evaluate the label aggregation strategies, we use the following label recovery accuracy computed on the evaluation datasets:
\begin{equation*}
    \mathrm{Acc}(\hat y, y; \mathcal{D}_{\bigcdot}) = \frac{1}{n_{\bigcdot}} \sum_{i=1}^{n_{\bigcdot}} \mathds{1}(\hat{y}_i = y_i)\mathds{1}(s_i=1) \enspace,
\end{equation*}
with $\hat y=(\hat{y}_i)_i$ the estimated labels on $\mathcal{D}_{\bigcdot} \in \{\mathcal{D}_\text{expert}, \mathcal{D}_{\text{multiple votes}}, \mathcal{D}_{\text{disagreement}}\}$, $y=(y_i)_i$ the associated experts labels, considered as ground truth.
When the aggregation strategy indicates the observation as invalid ($s_i=0$ for Pl@ntNet and TwoThird), this metric considers the sample as wrongly classified.
Precision and recall scores are also computed to respectively measure the correctness of the observations indicated as valid and the ability to recover the ground truth observations in the valid set.
We take into account the species imbalance by using a macro-average for these metrics. This treats rare species as equally important to common ones. Denoting respectively $\mathrm{TP}_k$, $\mathrm{FP}_k$ and $\mathrm{FN}_k$ the true positives, false positives and false negatives related to the species $k$, the macro averaged precision and recall write
\[
\mathrm{Precision}_{\mathrm{macro}}=\frac{1}{K}\sum_{k=1}^K \frac{\mathrm{TP}_k}{\mathrm{TP}_k + \mathrm{FP}_k} \quad \text{and} \quad \mathrm{Recall}_{\mathrm{macro}}=\frac{1}{K}\sum_{k=1}^K \frac{\mathrm{TP}_k}{\mathrm{TP}_k + \mathrm{FN}_k} \enspace.
\]

As both the Pl@ntNet and the TwoThird strategies can invalidate some of the observations, we also compute the proportion of observations removed from the whole dataset (whereas previous metrics are computed on the evaluation dataset). This complementary metric allows measuring the proportion of samples "lost" for the training of the AI model after the aggregation step.
In practice, filtering data might remove some noisy samples from the dataset. Yet, in general, the more samples are filtered, the fewer ones to train the neural network training.
Finally, we also consider the proportion of species retrieved by the aggregation strategies on $\mathcal{D}_\text{expert}, \mathcal{D}_{\text{mulitple votes}}$ and $\mathcal{D}_{\text{disagreement}}$. This is a critical consideration because if a species identified by experts is absent from the aggregated data, the neural network trained on this data will be unable to make predictions for that very species.

We evaluate the label recovery $\mathrm{Acc}$ of each strategy on $\mathcal{D}_\text{expert}, \mathcal{D}_{\text{mulitple votes}}$ and $\mathcal{D}_{\text{disagreement}}$ (see also \Cref{fig:sankey}): the test set where experts have provided at least one vote ($\mathcal{D}_\text{expert}$), its subset of observations with at least $2$ votes and one from an expert ($\mathcal{D}_\text{multiple votes}$) and its subset of observations with at least $2$ votes, one from an expert, and one disagreement ($\mathcal{D}_\text{disagreement}$).
The latter is the most challenging as it contains the observations with the most ambiguity.
We selected these subsets to investigate the label aggregation strategies' performance depending on the ambiguity level.

%%%%%%%%%%%%%%%%%%%%%%%%%%%%%%%%%%%%%%%%%%%%%%%%%%%%%%%%%%%%%%%%%%%%%%%%%%%%%%%
\subsection{Taking into account AI votes}
\label{subsec:ai_votes}
%%%%%%%%%%%%%%%%%%%%%%%%%%%%%%%%%%%%%%%%%%%%%%%g%%%%%%%%%%%%%%%%%%%%%%%%%%%%%%%%

While we restricted ourselves to the SWE subset, Pl@ntNet's data is collected internationally.
The more correctly identified observations are added to the training set, the better the prediction of the trained model for end-users.
This classifier is trained from valid observations and aggregated labels (see \Cref{fig:plantnet-system}).
% After the label aggregation step, the training is perform on valid observations to perform species classification (see \Cref{fig:plantnet-system}).
Note that, in addition to \Cref{alg:plantnet_algorithm} and the filter on species names, more pre-processing are implemented for better performance \citep{affouard2017pl}, such as additional rejection class (\emph{e.g.} non-plant observations), malformed observations (multiple images of different species in a single observation).
% Depending on the label aggregation strategy, using citizen annotations helps robustifying the model \citep{peterson_human_2019}.
At the time of writing, the model in use in Pl@ntNet is DINOv2 \citep{oquab2024dinov2} a transformer-based network.
This network is based on contrastive learning \citep{waida2023understanding}, and
represents similar images as close embedding to learn similar features for similar observations and then uses supervised learning to fine tune the model.
Several transformations are performed during training such as data augmentation \citep{yang2023image}, data standardization and label smoothing \citep{szegedy2016rethinking}.
However, note that some observations from $\mathcal{D}_{\text{SWE}}$ have been processed by an earlier version of Pl@ntNet's AI: either an InceptionV3 \citep{szegedy2015rethinking} or a BEIT \citep{bao2021beit} classifier. 
We can use the classifiers to generate votes.
For an observation $i$, the AI vote is denoted $y_i^{\text{AI}}\in[K]$. The probability output in the classifier's predicted species is denoted $\mathbb{P}(y_i^{\text{AI}})$.

If we consider the trained model as any other user, denoted as \textbf{AI as user}, the same label aggregation strategies as in \Cref{subsec:label_aggregation_strategy} are available.
However, with the Pl@ntNet aggregation algorithm, the AI weight increases drastically and overpowers human users (see \Cref{subsec:label_aggregation_performance_comparison}).
This would mean the next Pl@ntNet model is mostly trained on the predictions of the previous one.
This defeats the purpose of a cooperative active learning system and the human-AI interaction. It would result in a dangerous feedback loop, and possible mode collapse.
Thus, we explore alternative ways of integrating the AI votes in the aggregation algorithm:
\begin{itemize}
    \item \textbf{AI as user}: This is the naive approach we just described. The AI is considered as any other user in the database. The total number of users is thus raised to $n_{\text{user}}+1$.
    \item \textbf{Fixed weight AI}: Give a fixed weight $w_{\text{AI}}=1.7>0$ to the \text{AI}. The weight is below the threshold $\theta_{\text{conf}}$ so that it can not self-validate its predictions. The confidence writes 
    \begin{equation}\label{eq:confweight}\mathrm{conf}_i(\hat y_i) = \sum_{u\in\mathcal{U}_i} w_u\mathds{1}(y_i^u=\hat y_i)+ w_{\text{AI}}\mathds{1}(y_i^{\text{AI}}=k)\enspace.
    \end{equation} The final estimated label becomes
    \begin{equation}\label{eq:wmvAI}\hat y_i = \argmax_{k\in [K]} \sum_{u\in \mathcal{U}_i} w_u \mathds{1}(y_i^u=k) + w_{\text{AI}}\mathds{1}(y_i^{\text{AI}}=k)\enspace.
    \end{equation}
    \item \textbf{Invalidating AI}: The \text{AI} is considered as a user with a fixed weight and can only participate in invalidating identifications \emph{i.e.} have $s_i=0$. This translates as the confidence updated as in \Cref{eq:confweight} but the final Weighted MV remains unchanged from \Cref{alg:plantnet_algorithm}.
    \item \textbf{Confident AI}: The \text{AI} is considered a user with a fixed weight and can only participate if the confidence in its prediction $\mathbb{P}(y_i^{\text{AI}})$ is over a threshold $\theta_{\text{score}}\in [0,1]$. 
    The confidence writes 
    \begin{equation}\mathrm{conf}_i(\hat y_i) = \sum_{u\in\mathcal{U}_i} w_u\mathds{1}(y_i^u=\hat y_i) + w_{\text{AI}}\mathds{1}(y_i^u=\hat y_i,\,\mathbb{P}(y_i^{\text{AI}})\geq\theta_{\text{score}})\enspace. \end{equation} The final estimated label becomes
\begin{equation}\hat y_i = \argmax_{k\in [K]} \sum_{u\in \mathcal{U}_i} w_u \mathds{1}(y_i^u=k) + w_{\text{AI}}\mathds{1}(y_i^{\text{AI}}=k,\,\mathbb{P}(y_i^{\text{AI}})\geq\theta_{\text{score}})\enspace.\end{equation}
\end{itemize}

\paragraph{On the choice of the AI weight.} The AI has a fixed weight $w_\text{AI}>0$ for the \textbf{Fixed weight AI}, the \textbf{Invalidating AI} and the \textbf{Confident AI} strategies.
The choice of this weight must meet several constraints.  
First, we would like to avoid the \text{AI} votes to be \textbf{self-validating} as it would validate all the \text{AI} predictions on a large part of the database, thus we must have $w_\text{AI}<\theta_\mathrm{conf}$ in \Cref{alg:plantnet_algorithm}.
We also want the \text{AI} votes to help clean the database by invalidating some observations from low-weight users (with weight 0< $w_\text{low} \leq \theta_\mathrm{conf}$). Thus $w_\text{low} / (w_\text{low} + w_\text{AI}) < \theta_\mathrm{acc}$.
Hence, our constraints read:
\begin{equation}\label{eq:constraint_weight}
\begin{cases} w_\text{AI} &< \theta_\mathrm{conf} \\ \frac{w_\text{low}}{w_\text{low} + w_\text{AI}} &< \theta_\mathrm{acc} \end{cases} \enspace.
\end{equation}
Taking the extreme case where a user becomes self-validating: $w_\text{low}=\theta_\mathrm{conf}$, we obtain that $w_\text{AI} > \theta_\mathrm{conf} \left(\frac{1-\theta_\mathrm{acc}}{\theta_\mathrm{acc}}\right)$. And using the first condition in \Cref{eq:constraint_weight}, we obtain the bounds
\begin{equation}\label{eq:constraint_numerical}
\theta_\mathrm{conf} \left(\frac{1-\theta_\mathrm{acc}}{\theta_\mathrm{acc}}\right) < w_\mathrm{AI} < \theta_\mathrm{conf} \left(\Longleftrightarrow 0.85 < w_\mathrm{AI} < 2 \right)\enspace.
\end{equation}
As more than a million observations from our dataset only have two votes, one way to choose the \text{AI} weight is to consider that the \text{AI} can invalidate two erroneous non-experts that would both have just enough weights to make the observation valid: $1.95=w_\text{low} < 2$.
Then, the AI weight should be greater than their cumulated confidence: $w_\mathrm{AI}>2w_\text{low}\left(\frac{1-\theta_\mathrm{acc}}{\theta_\mathrm{acc}}\right)$. We finally take the upper rounded value $w_\mathrm{AI}=1.70$ (which satisfies \Cref{eq:constraint_numerical}).

% \begin{figure}[thb]
%         \begin{center}
%             \includegraphics[width=0.85\textwidth]{./images/distribution_shift.pdf}
%         \end{center}
%         \caption{Distribution of scores for the predicted class by the AI model depending on the level of ambiguity of the observation. The more ambiguity there is, the more bimodal the distribution is and the less confident the model is.}
%         \label{fig:distrib_shift}
% \end{figure}

\section{Results}

%%%%%%%%%%%%%%%%%%%%%%%%%%%%%%%%%%%%%%%%%%%%%%%%%%%%%%%%%%%%%%%%%%%%%%%%%%%%%%%

%%%%%%%%%%%%%%%%%%%%%%%%%%%%%%%%%%%%%%%%%%%%%%%%%%%%%%%%%%%%%%%%%%%%%%%%%%%%%%%
\subsection{Label aggregation performance comparison}
\label{subsec:label_aggregation_performance_comparison}
%%%%%%%%%%%%%%%%%%%%%%%%%%%%%%%%%%%%%%%%%%%%%%%%%%%%%%%%%%%%%%%%%%%%%%%%%%%%%%%

\begin{figure}[tbh]
    \centering
    \begin{subfigure}[t]{.48\textwidth}
        \centering
        \includegraphics[width=\linewidth, height=4cm]{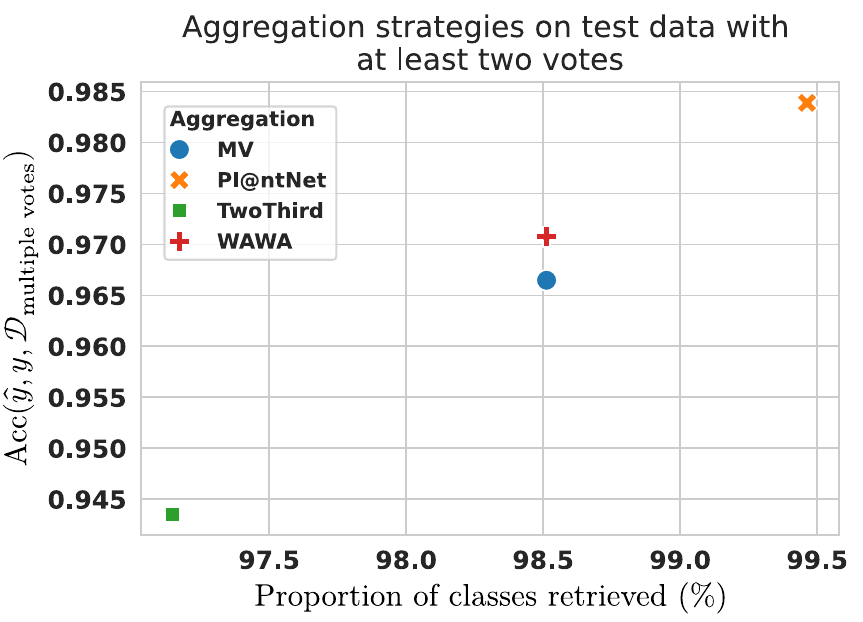}
        \caption{Accuracy on $\mathcal{D}_{\text{multiple votes}}$ w.r.t. to the proportion of classes recovered}
        \label{fig:accuracy_vol_class_multiple}
    \end{subfigure}%
    \hfill
    \begin{subfigure}[t]{.48\textwidth}
        \centering
        \includegraphics[width=\linewidth, height=4cm]{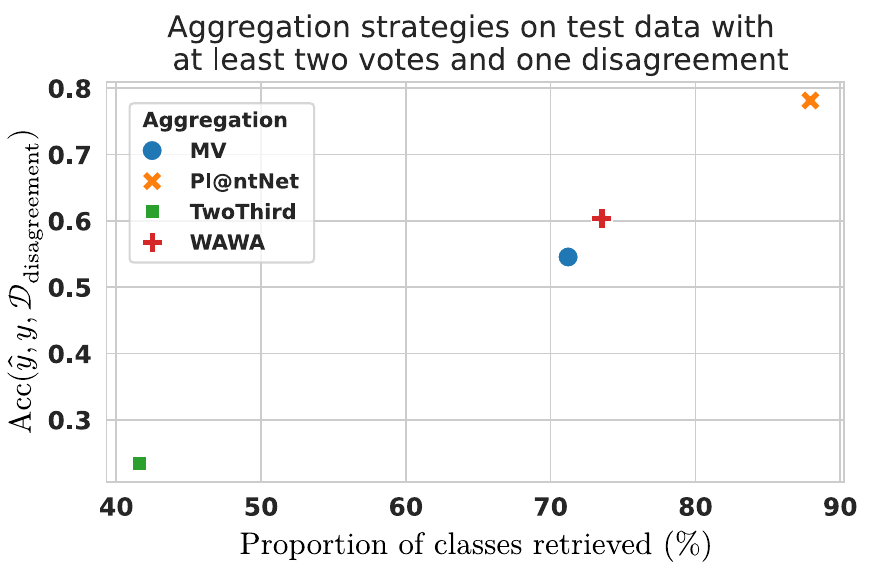}
        \caption{Accuracy on $\mathcal{D}_{\text{disagreement}}$ w.r.t. the proportion of classes recovered}
        \label{fig:accuracy_vol_class_disagreeing}
    \end{subfigure}
    \caption{Accuracy of the aggregation strategies w.r.t. the proportion of classes (species) retrieved on subsets with at least two votes -- either agreeing (A) or with at least one disagreeing vote (B). The Pl@ntNet aggregation is more accurate, especially in a highly ambiguous setting (B). The TwoThird data filter highly impacts how many classes are kept in the dataset and the overall accuracy in both settings. WAWA and MV perform similarly with a benefit for WAWA when skill evaluation is needed.}
    \label{fig:accuracy_volume}
\end{figure}

\paragraph*{Accuracy of the aggregation strategies.} 

We begin by evaluating the accuracy of the label aggregation strategies on the set of observations labeled by experts, $\mathcal{D}_\text{expert}$. \Cref{fig:accuracy_volume} shows how many predicted labels match the experts answers on $\mathcal{D}_{\text{multiple votes}}$ and $\mathcal{D}_{\text{disagreement}}$.
More importantly, we compare this quantity with the proportion of species retrieved by the aggregation strategy.
We observe that the data filtering from the TwoThird strategy -- requiring at least two third of agreements -- highly degrades performance with respect to other strategies.
On $\mathcal{D}_{\text{expert}}$, MV reaches $97\%$ of accuracy, WAWA $98\%$, TwoThird $60\%$ and Pl@ntNet $99\%$.
To differentiate between the best-performing strategies, we need to look at more ambiguous observations like those in $\mathcal{D}_{\text{multiple votes}}$ and $\mathcal{D}_{\text{disagreement}}$.
In highly ambiguous frameworks, the WAWA strategy outperforms the MV one.
However, overall the Pl@ntNet aggregation is more often in adequation with the experts and retrieves almost $90\%$ of plant species identified by experts in highly ambiguous datasets against $73\%$ for WAWA, $71\%$ for MV and only $41\%$ for TwoThird.

\begin{figure}[tbh]
    \centering
    \begin{subfigure}[t]{.48\textwidth}
        \centering
    \includegraphics[width=\textwidth]{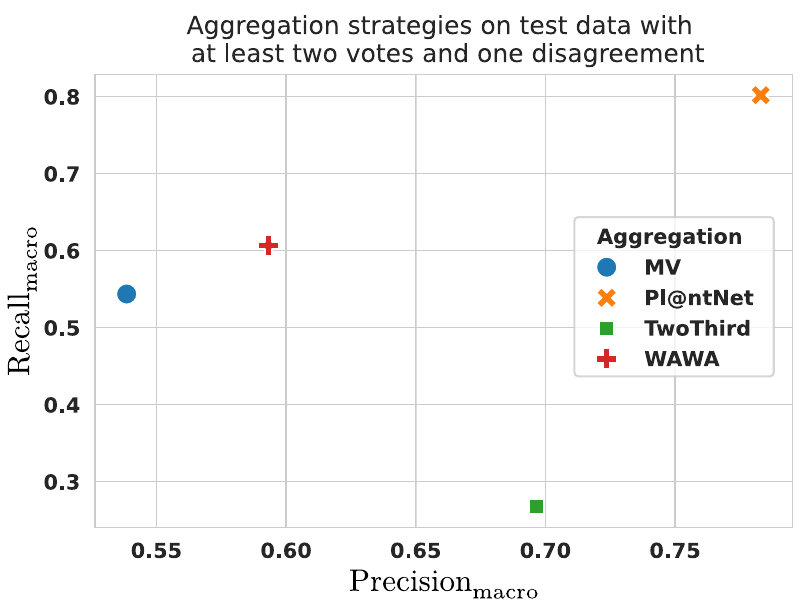}
    \caption{Precision and recall of label aggregation strategies on $\mathcal{D}_\text{disagreement}$. }
        \label{fig:precision-recall}
    \end{subfigure}%
    \hfill
    \begin{subfigure}[t]{.48\textwidth}
        \centering
    \includegraphics[width=\textwidth]{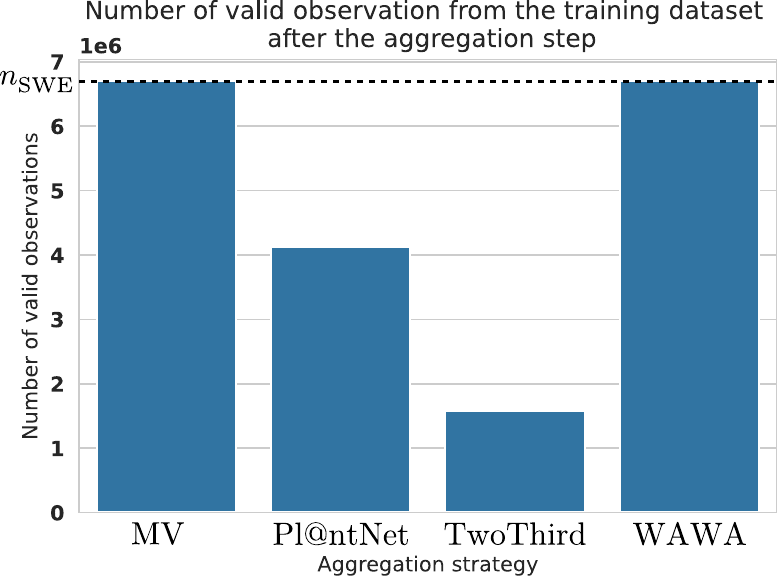}
    \caption{Number of observations in $\mathcal{D}_{\mathrm{SWE}}$ indicated as valid for training ($s_i=1$). }
    \label{fig:vol_train_data}    \end{subfigure}
    \caption{(A): TwoThird strategy has better precision than MV and WAWA strategies, with lower recall because of the heavy filter on the validity of observations. Pl@ntNet aggregation strategy obtains best precision and recall and outperforms other strategies. (B): TwoThird performance drop can be explained in part by the high proportion of data considered invalid. Note that MV and WAWA strategies do not invalidate any observation, hence keeping potentially mislabeled or low-quality observations. Pl@ntNet achieves a balance between filtering out observations and achieving high performance.}
    \label{fig:precision-volume}
\end{figure}

\begin{figure}[tbh]
    \centering
    \includegraphics[width=.85\textwidth]{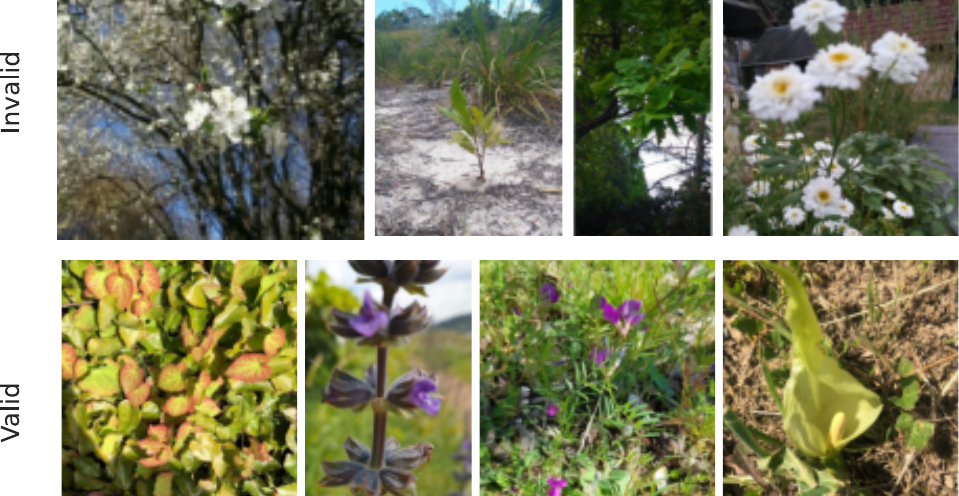}
    \caption{Examples of invalid $(s_i=0)$ and valid ($s_i=1$) observations using the Pl@ntNet strategy described in \Cref{alg:plantnet_algorithm}.}
    \label{fig:valid-invalid-plantnet}
\end{figure}

\paragraph*{Precision and recall.}

To better evaluate each aggregation strategy, we compute the macro precision and recall metrics for each species. Results are shown in \Cref{fig:precision-recall}. The observations filter ($s_i=0$) for the TwoThird strategy highly impacts its ability to identify most of the positive observations for a given species. While this agreement threshold filter is created to keep as few noisy samples as possible in research-graded (data quality indicator for research database usage in TwoThird) observations, TwoThird obtains better precision than MV and WAWA but Pl@ntNet's precision shows significant improvement. WAWA strategy outperforms a naive MV aggregation showing that, indeed, weighing users can lead to better performance. Pl@ntNet strategy outperforms all others by several orders of magnitude. Weighing users based on their number of identified species is both interpretable and effective. The observation filter does not negatively impact the recall.

\paragraph*{Volume of valid data.}

The community labels are aggregated to generate training data for the AI model.
The more data the better, however, we need to filter out observations with low visual quality or potentially mislabeled.
This is the reason for the validity indicator $s_i$ in the TwoThird and Pl@ntNet strategies.
On $\mathcal{D}_{\mathrm{SWE}}$, \Cref{fig:vol_train_data} shows how much data is kept for later training.
$\mathrm{MV}$ and $\mathrm{WAWA}$ keep all proposed observation for training -- including potential noisy ones.
TwoThird filters out most observations to keep nearly $1.5$ million (representing $23.43\%$ of the total observations). Pl@ntNet finds an improved balance between filtering invalid observations and keeping enough data for training. 

\paragraph*{Qualitative results on Pl@ntNet observation filter.}

In this section, we show some examples of observations invalidated by the Pl@ntNet strategy (see \Cref{fig:valid-invalid-plantnet}).
Invalid observations often come from the lack of user participation with other's observations.
Causes of disagreements from users can occur from a multitude of factors -- blurriness, multiple species in the same observation, the distance from the plant does not allow precise identification, \emph{etc.}
Valid observations, as shown in the second row of \Cref{fig:valid-invalid-plantnet} are zoomed in on the plant's flower, leaf or organ to help the identification process.

%%%%%%%%%%%%%%%%%%%%%%%%%%%%%%%%%%%%%%%%%%%%%%%%%%%%%%%%%%%%%%%%%%%%%%%%%%%%%%%
\subsection{Aggregation considering AI vote}
\label{subsec:aggregation_and_ai_vote}
%%%%%%%%%%%%%%%%%%%%%%%%%%%%%%%%%%%%%%%%%%%%%%%%%%%%%%%%%%%%%%%%%%%%%%%%%%%%%%%

\begin{figure}[tbh]
    \centering
    \includegraphics[width=\textwidth]{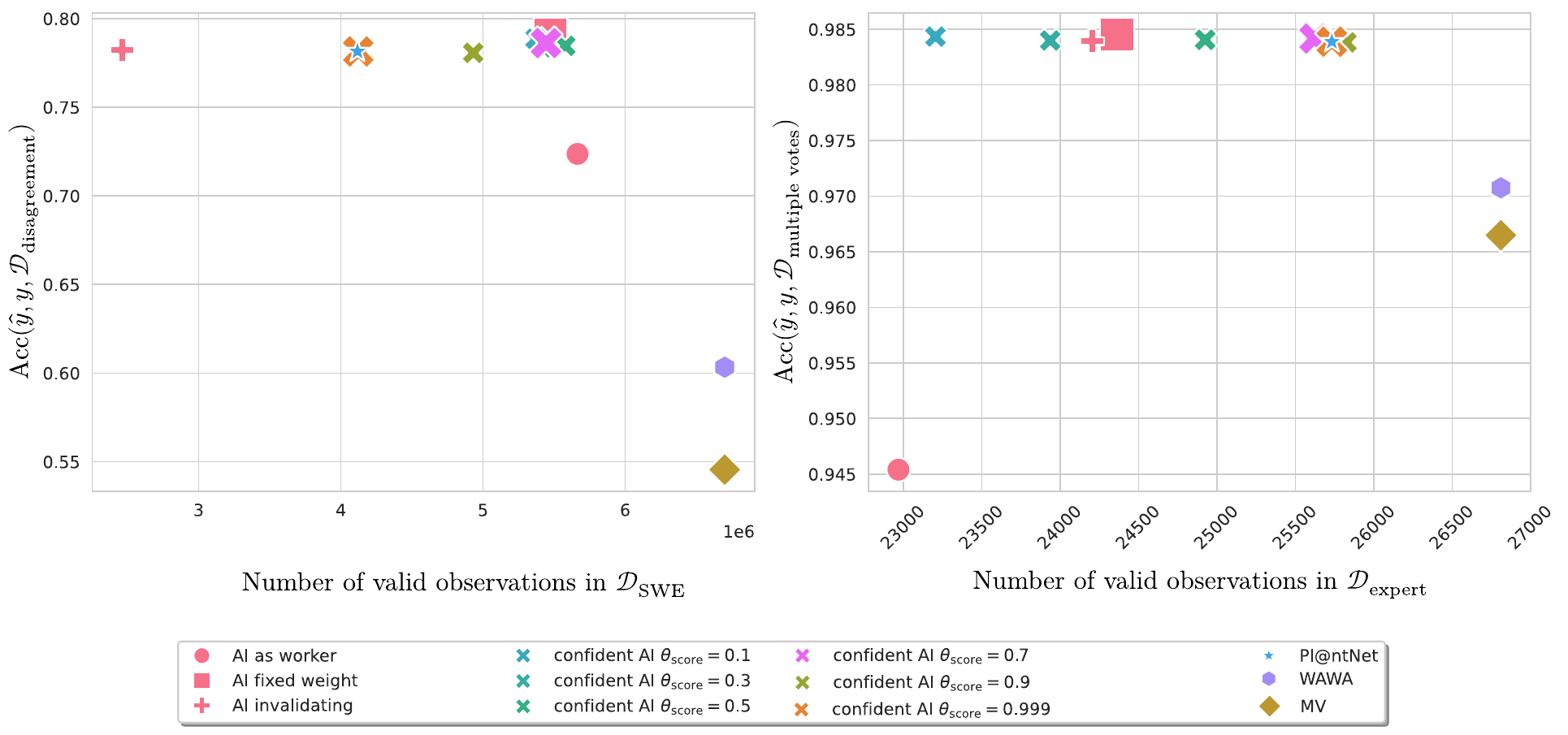}
    \caption{Performance in label recovery and number of observations marked as valid depending on how the AI vote is integrated. MV, WAWA and Pl@ntNet strategy without AI vote are used as reference. The best-performing strategy overall is \textbf{confident AI} with $\theta_\text{score}=0.7$. We also see that when $\theta_\text{score}$ tends to $1$, we recover the vanilla Pl@ntNet aggregation strategy.}
\label{fig:accuracy_vote_strategy}
\end{figure}

The current trained neural network model in Pl@ntNet's system can make predictions based on its training on the Pl@ntNet database (across different floras).
We compare the four following strategies -- \textbf{AI as user}, \textbf{fixed weight AI}, \textbf{invalidating AI} and \textbf{confident AI}, presented in \Cref{subsec:ai_votes} to integrate the AI vote into the Pl@ntNet label aggregation strategy.
For the confident AI strategy, we evaluate multiple thresholds $\theta_\text{score}$.
Note that if $\theta_{\mathrm{score}}=0$ the AI votes for all observations and if $\theta_{\mathrm{score}}=1$ the AI does not vote and we recover the performance of the current Pl@ntNet aggregation strategy presented in \Cref{alg:plantnet_algorithm}.
We see in \Cref{fig:accuracy_vote_strategy} that the confident AI strategy with $\theta_\mathrm{score}=0.7$ seems to perform best and keep the most data in both $\mathcal{D}_\mathrm{SWE}$ and $\mathcal{D}_\mathrm{expert}$.
% However, we notice that the less confident the AI is in its prediction, the more likely the prediction does not match the expert's label (see \Cref{tab:experts_and_AI}).
% Hence why we take into account in the \textbf{Confident AI} strategy its prediction score.

% \begin{table}[thb]
%     \caption{Accuracy (in $\%$) of the current AI model on our test set depending on the confidence score of the prediction. The less confident, the less accurate the prediction.}
%     \label{tab:experts_and_AI}
%     \centering
%     \begin{tabular}{lccc}
%         \hline
%         Subset & All observations & Score $\geq 80\%$ & Score $< 80\%$  \\
%         \hline
%         More than two votes & $85.28$ & $96.84$ & $75.37$ \\
%         One disagreement & $64.45$ & $82.69$ & $54.73$ \\
%         \hline
%     \end{tabular}
% \end{table}

\section{Discussion}

We demonstrated that collaborative identification of plant species can effectively be used to obtain expert level labels.
Releasing a large subset of millions of observations and thousands of users from the Pl@ntNet organization, we investigate a label aggregation strategy that weighs user answers based on their estimated number of species correctly identified without using prior expert knowledge.
Many strategies used previously either do not scale to the magnitude of the current databases -- either Pl@ntNet, iNaturalist or eBird -- or are outperformed by our aggregation.

Our strategy weighs users based on the number of correctly identified species. This weight is interpretable and shows the diversity of the user's skill set.
It can be directly applied to other crowdsourced frameworks with a high number of classes like TwoThird or eBird.
The values for both hyperparameters $\theta_\mathrm{conf}$ and $\theta_\mathrm{acc}$ which respectively handle the cumulated weight on observation and the agreement level for the given label can be applied as is.

Note that Pl@ntNet's label control system heavily rests on visual analysis of observations and inter-user agreements. Additional metadata such as geolocation, date, phenological stage or visual description can be registered in Pl@ntNet and help identify the plant's species but are currently not directly taken into account for user evaluation. Such information -- in particular spatial information -- could also be used to generate more interaction between users and collect more votes through possible common interests.
In addition, users are helped by the system with images similar to the identification proposed in a given checklist.
The additional information could guide users in their vote -- for example by notifying a possible incoherence between the current botanical knowledge on a species and the metadata entered (such as the altitude, the distance to the sea, a species not known to survive in a given area).

As for the inclusion of the AI vote, some concerns should be raised.
First, as the AI model is trained from the aggregated labels and observations, integrating its vote should not make the AI predictions run out of control.
If we consider the AI as a user, as we are in iterative training, the system fails to learn from the human labels.
However, using the AI vote to invalidate the data with a fixed weight can help clean the database, and with enough weight other users can switch its validity back. However, this would not help in switching the wrong label.
To do so, we investigate in \Cref{subsec:aggregation_and_ai_vote} to only consider a fixed weight label with enough confidence from the AI model.
We observe that this strategy leads to better performance.
As we use the output probabilities we should discuss the calibration of our network too.

\begin{figure}[tbh]
    \centering
    \begin{subfigure}[t]{.48\textwidth}
        \centering
    \includegraphics[width=\textwidth]{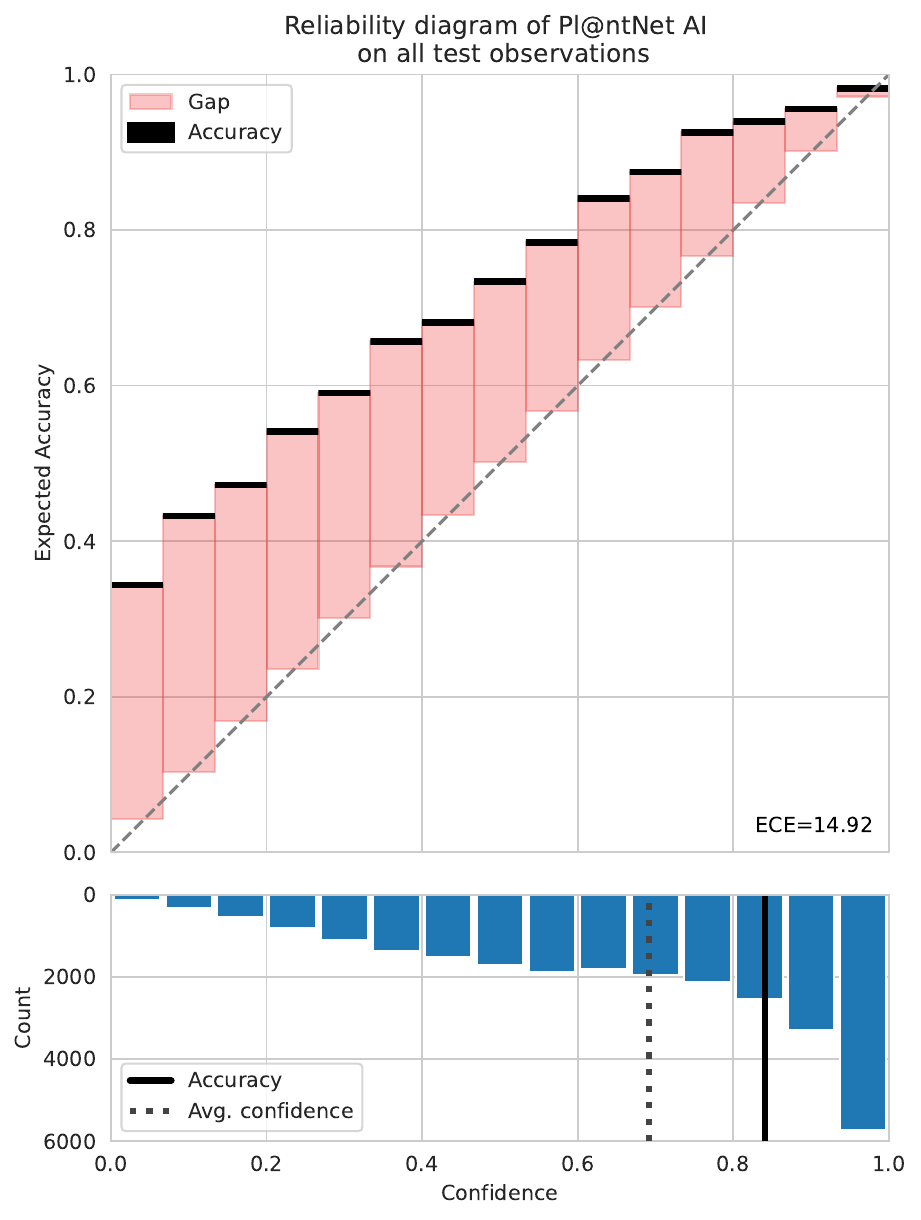}
    \caption{Reliability diagram of Pl@ntNet AI on $\mathcal{D}_{\mathrm{expert}}$ }
        \label{fig:reliability-all}
    \end{subfigure}%
    \hfill
    \begin{subfigure}[t]{.48\textwidth}
        \centering
    \includegraphics[width=\textwidth]{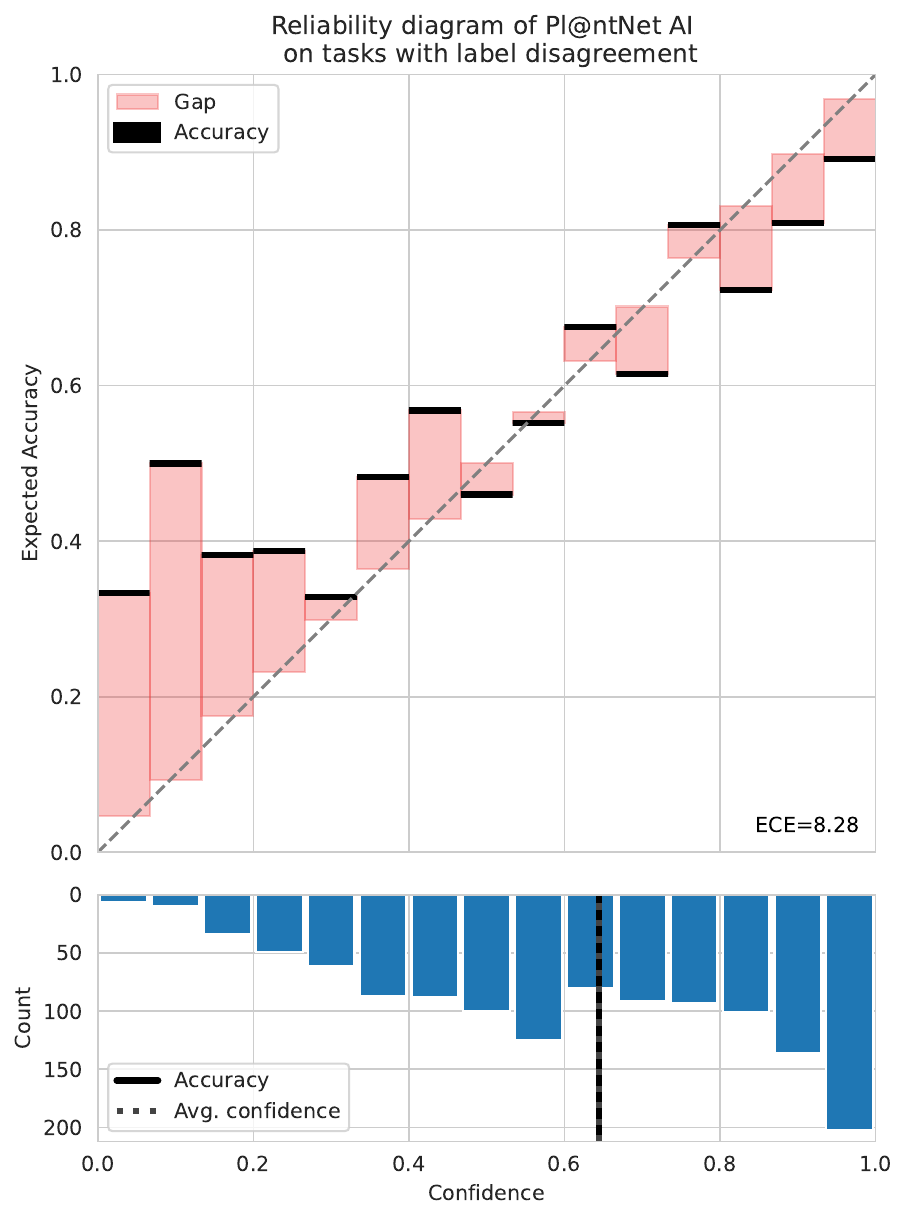}
    \caption{Reliability diagram of Pl@ntNet AI on $\mathcal{D}_{\mathrm{disagreement}}$ }
    \label{fig:reliability-disagreement}    \end{subfigure}
    \caption{Reliability diagrams for the Pl@ntNet AI on the expert dataset and on the more ambiguous $\mathcal{D}_\mathrm{disagreement}$ subset. The AI is overall underconfident (A). However, on more ambiguous observations it is overconfident for observations leading to high predicted probabilities (B).}
    \label{fig:reliability}
\end{figure}

Calibration is the measure of how close the output confidence is to the true probability \citep{niculescu2005predicting}. Currently, the Pl@ntNet AI is not calibrated using post-processing methods \citep{platt1999probabilistic, guo_calibration_2017}.
We discuss hereafter the calibration of the current AI model and possible guidelines for further integration of AI votes.

From \Cref{fig:reliability-all}, we see that currently, Pl@ntNet AI is underconfident. Meaning that it consistently underestimates its confidence and outputs to users more uncertainty than it should.
One factor that can influence the results is that the calibration is computed on the test set where experts either authored or voted on observations.
Botanical experts have more experience with taking pictures of plants and better equipment than the average citizen.
Thus, the observation quality -- and subsequently the probability distribution output by the AI -- can be biased.
Another factor known for leading to such suboptimal predictions is the data augmentation \citep{kapoor2022uncertainty}. As the model trains on multiple versions of each original sample with multiple distortions, these variations can become unrepresentative of the underlying sample distribution and cause unnecessary prediction difficulties.
The data augmentation is used to mitigate the species imbalance of the database.

However, this imbalance is also known to lead to miscalibrations in predictions \citep{ao2023two}.
On \Cref{fig:reliability-disagreement}, we see that for ambiguous observations (where users disagree), the AI is overconfident in its highest predictions -- which represents half of the dataset -- and underconfident in the other half.
These different calibration behaviors inform us that, if a given strategy should incorporate the AI votes in the label aggregation based on the output probabilities, we need to be able to rely on such probabilities.
Therefore, even if the confident AI strategy leads to the best performance in \Cref{subsec:ai_votes}, it should not be used directly without recalibration of the model -- using for example temperature scaling \citep{guo_calibration_2017}.
In future work, more study is needed to investigate the confidence gap of the model and the observations' ambiguity from users' labels.
The current large-scaled and interpretable aggregation strategy from Pl@ntNet already outperforms others without the AI votes.

% \section*{Acknowledgments}
% This work was funded by the French National Research Agency (ANR) through the grant Pl@ntAgroEco 22-PEAE0009, granted access to the HPC resources of IDRIS under the allocation A0151011389 made by GENCI, and funded by the Chaire IA CaMeLOt (ANR-20-CHIA-0001-01).

% \section*{Conflict of interest statement}
% The authors declare no conflicts of interest.

% \section*{Author contributions}
% Tanguy Lefort, Antoine Affouard, Alexis Joly, Benjamin Charlier, Pierre Bonnet and Joseph Salmon conceived the ideas and designed the evaluation methodology; Antoine Affouard and Mathias Chouet are the main developers of Pl@ntNet's backend ; Tanguy Lefort and Antoine Affouard collected the evaluation data used in this paper; Tanguy Lefort re-implemented Pl@ntNet's algorithm in python and conducted the evaluation ; Jean-Christophe Lombardo, Hervé Goëau and Alexis Joly conceived and trained Pl@ntNet's AI model; Tanguy Lefort, Benjamin Charlier, Alexis Joly and Joseph Salmon analyzed the outcomes of the study; Tanguy Lefort, Benjamin Charlier, Alexis Joly and Joseph Salmon led the writing of the manuscript. All authors contributed critically to the drafts and gave final approval for publication.

\section*{Statement on inclusion}
We affirm our commitment to promoting diversity and inclusivity in scientific research. Our collected crowdsourced data brings together a wide range of participants. We actively encourage and welcome involvement from individuals of diverse backgrounds, expertise, and perspectives, recognizing the value of their contributions in advancing ecological research and promoting a more comprehensive understanding of plant biodiversity.

% \section*{Data availability}
% The dataset is available at \url{https://doi.org/10.5281/zenodo.10782465}

\printbibliography
\end{document}